\documentclass[11pt,a4paper]{article}
\usepackage[hyperref]{acl2020}
\usepackage{times}
\usepackage{latexsym}

\usepackage{color}
\usepackage{tabularx}
\usepackage{microtype}
\usepackage{amsfonts}
\usepackage{amssymb}
\usepackage{amsmath}
\usepackage{mathrsfs}
\usepackage{multirow}
\usepackage{array}
\usepackage{graphicx}
\usepackage{subfigure}
\usepackage{float}
\usepackage{algorithm}
\usepackage{algorithmic}
\usepackage{xcolor}
\usepackage{colortbl,booktabs}
\aclfinalcopy 
\def\aclpaperid{623} 

\title{Conditional Augmentation for Aspect Term Extraction via Masked Sequence-to-Sequence Generation}

\author{Kun Li\textsuperscript{1}, Chengbo Chen\textsuperscript{1}, Xiaojun Quan\textsuperscript{1}\thanks{\; Corresponding author.}, Qing Ling\textsuperscript{1}, Yan Song\textsuperscript{2}\\
\textsuperscript{1}School of Data and Computer Science, Sun Yat-sen University, China \\
\textsuperscript{2}Sinovation Ventures, China\\
\tt \{likun36, chenchb7\}@mail2.sysu.edu.cn \\
\tt \{quanxj3, lingqing556\}@mail.sysu.edu.cn, \tt clksong@gmail.com}

\date{}

\begin{document}
\maketitle
\begin{abstract}
Aspect term extraction aims to extract aspect terms from review texts as opinion targets for sentiment analysis. One of the big challenges with this task is the lack of sufficient annotated data.~While data augmentation is potentially an effective technique to address the above issue, it is uncontrollable as it may change aspect words and aspect labels unexpectedly. In this paper, we formulate the data augmentation as a conditional generation task: generating a new sentence while preserving the original opinion targets and labels.~We propose a masked sequence-to-sequence method for conditional augmentation of aspect term extraction.~Unlike existing augmentation approaches, ours is controllable and allows us to generate more diversified sentences.~Experimental results confirm that our method alleviates the data scarcity problem significantly. It also effectively boosts the performances of several current models for aspect term extraction.

\end{abstract}

\section{Introduction}
\label{introduction}
Aspect term extraction (ATE), which aims to identify and extract the aspects on which users express their sentiments \cite{hu2004mining,liu2012sentiment}, is a fundamental task in aspect-level sentiment analysis. 
For example, in the sentence of “\textit{The screen is very large and crystal clear with amazing colors and resolution}”, “\textit{screen}”, “\textit{colors}” and “\textit{resolution}” are the aspect terms to extract in this task.

\begin{figure}[h]
\centering
\includegraphics[width=1\linewidth]{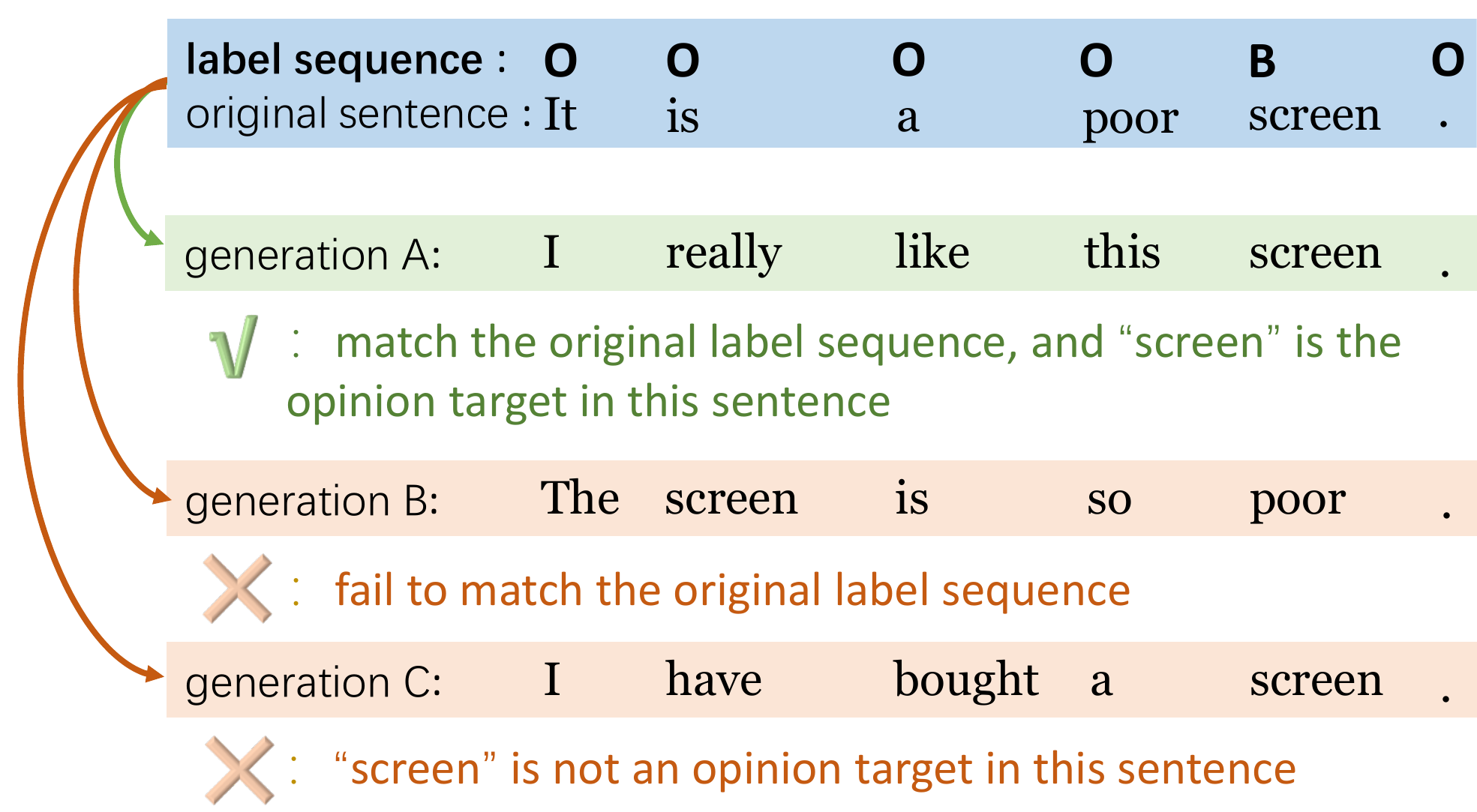}
\caption{Examples of ATE augmentation, where \emph{B}, \emph{I} and \emph{O} denote that a word is the beginning, inside and outside of opinion target, respectively.}
\label{example}
\end{figure}

ATE is typically formulated as a sequence labeling problem \cite{xu2018double,xu-etal-2019-bert,li2018aspect}, where each word is appended with a label indicating if it identifies an aspect. Sentence and label sequence are both used to train a ATE model. One of the remaining challenges with this task is the shortage of annotated data. While data augmentation appears to be a solution to this problem, it faces two main obstacles here. First, the new sentences must adhere to their original label sequences strictly. As shown in Figure \ref{example}, the \emph{generation A} is an effective augmentation as the original label sequence is preserved, whereas \emph{B} is not even though it can be a valid review. Second, a noun phrase is regarded as aspect term only if it is an opinion target. In the \emph{generation D} of Figure \ref{example}, although the term ``\emph{screen}'' remains where it is in the original sentence, the new context makes it just an ordinary mention rather than an opinion target. To sum up, the real difficulty of data augmentation in ATE is generating a new sentence while aligning with the original label sequence and making the original aspect term remain an opinion target. Existing augmentation models such as GAN \cite{goodfellow2014generative} and VAE \cite{kingma2013auto} tend to change the opinion target unpredictably and thus are not applicable for this task. 

Another genre of augmentation strategy is based on word replacement. It generates a new sentence by replacing one or multiple words with their synonyms \cite{zhang2015character} or with words predicted by a language model \cite{kobayashi2018contextual}. This approach seems to be able to address the above issue in ATE augmentation, yet it only brings very limited changes to the original sentences and cannot produce diversified sentences.~Intuitively, augmentation strategies are effective when they increase the diversity of training data seen by a model.

We argue in this paper that the augmentation for aspect term extraction calls for a conditional approach, which is to be formulated as a masked sequence-to-sequence generation task. Specifically, we first mask several consecutive tokens for an input sentence. Then, our encoder takes the partially masked sentence and its label sequence as input, and our decoder tries to reconstruct the masked fragment based on the encoded context and label information. 
The process of reconstruction keeps the opinion target unchanged and is therefore controllable. Moreover, compared with replacement-based approaches \cite{zhang2015character,kobayashi2018contextual} which replace words separately, ours replaces a segment each time and has the potential to generate more diversified new sentences in content.

To implement the above conditional augmentation strategy, we adopt Transformer \cite{vaswani2017attention} as our basic architecture and train it like MASS \cite{song2019mass}, a pre-trained model for masked sequence-to-sequence generation. 

The contributions of this work are as follows. 
\begin{itemize}
    \item To our knowledge, this work is the first effort towards data augmentation of aspect term extraction through conditional text generation. 
    \item We propose a controllable data augmentation method by masked sequence-to-sequence generation, which is able to generate more diversified sentences than previous approaches. 
    \item We provide qualitative analysis and discussions as to why our augmentation method works, and test its implementation with other language models to illustrate why this masked sequence-to-sequence framework is favored.
\end{itemize}

\begin{figure*}[h]
\centering
\includegraphics[width=0.99\linewidth]{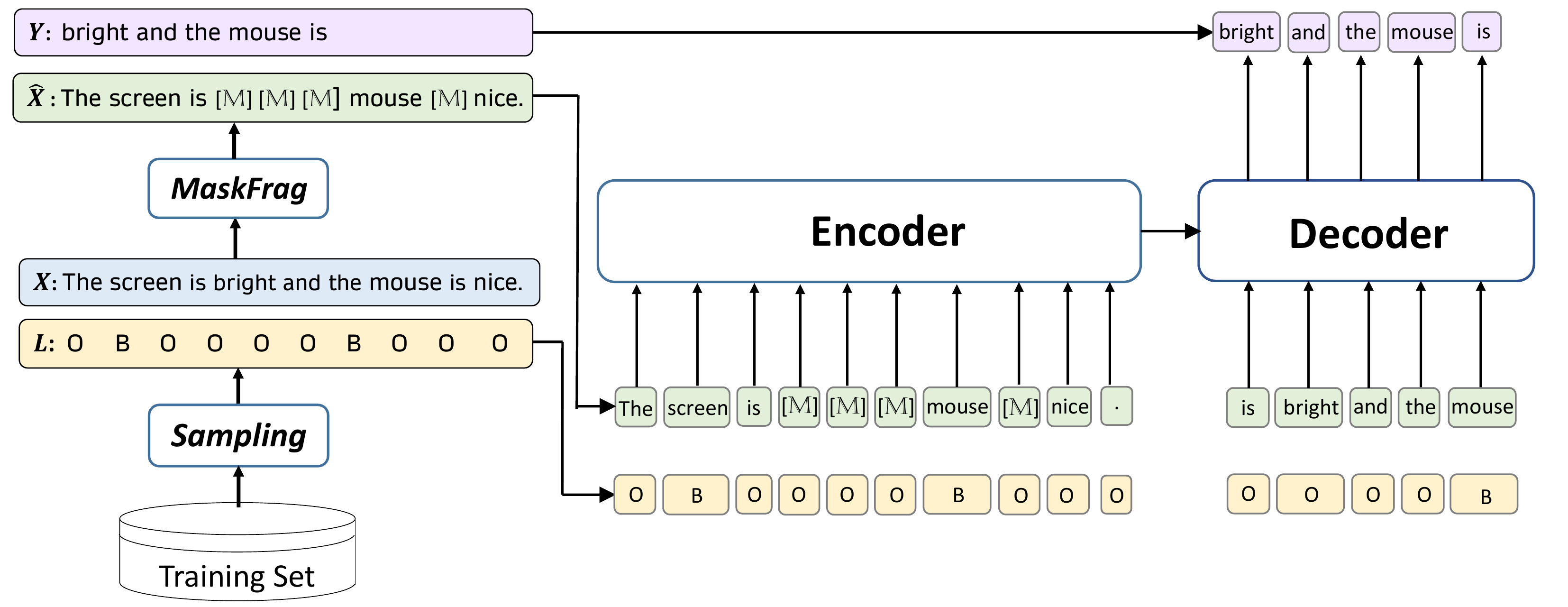}
\caption{Framework of our augmentation method.}
\label{framework}
\end{figure*}

\section{Related Work}

\subsection{Aspect Term Extraction}
Aspect term extraction (ATE) and sentiment classification are two fundamental subtasks of aspect-based sentiment analysis.~While the former aims to extract aspect terms in review sentences, the latter tries to determine their sentiment polarities.~To deal with ATE, many traditional techniques like syntactic rules \cite{qiu-etal-2011-opinion}, hidden Markov models \cite{jin2009novel}, and conditional random fields \cite{li2010structure,toh2016nlangp} have been explored.

Recently, neural network techniques such as LSTM \cite{liu2015fine}, CNN \cite{xu2018double}, and attention \cite{li2018aspect,devlin2019bert} have been applied for ATE. \citet{luo2019doer} and \citet{he2019interactive} further proposed to predict aspect term and polarity jointly in a multi-task learning approach so as to take advantage of their relatedness.~Generally, the above approaches treat ATE as a sequence labeling problem. In their pioneering work, \citet{ma2019exploring} formulated ATE as a sequence-to-sequence task. So far, one of the remaining challenges for ATE lies in the lack of annotated data, especially when today's neural models are becoming increasingly large and complex.

\subsection{Text Data Augmentation}
Generative adversarial network (GAN) \cite{goodfellow2014generative} and variational autoencoder (VAE) \cite{kingma2013auto} are two neural network based generative models that are capable of generating text conditioned on input text and can be applied for data augmentation of sentence-level sentiment analysis \cite{gupta2019data, Hu:2017:TCG:3305381.3305545}. 
These methods encode an input text into latent variables and generate new texts by decoding the latent variables in continuous space. However, they can hardly ensure high-quality sentences in terms of readability and label compatibility. Back translation \cite{edunov2018understanding,sennrich2015improving} is another augmentation approach for text data, but is less controllable, although it is good at maintaining the global semantics of an original sentence. As a class of replacement approach, \citet{zhang2015character} and \citet{wang2015s} proposed to substitute all replaceable words with corresponding synonyms from WordNet \cite{miller1995wordnet}. Differently, \citet{kobayashi2018contextual} and \citet{DBLP:conf/iccS/WuLZHH19} proposed to randomly replace words with those predicted by a pre-trained language model.

Nevertheless, none of the above augmentation approaches is applicable for aspect term extraction task, as they are all targeted at sentence-level classification and may change opinion targets and aspect labels unexpectedly during augmentation. 

\subsection{MASS}
Pre-training a large language model and fine-tuning it on downstream tasks has become a new paradigm. MASS \cite{song2019mass} is such a model for language generation. Unlike GPT \cite{radford2016improving,radford2019language} and BERT \cite{devlin2019bert} which only have either an encoder or a decoder, MASS includes both of them and trains them jointly: the encoder takes as input a sentence with a fragment masked and outputs a set of hidden states; the decoder estimates the probability of a token in the masked fragment conditioned on its preceding tokens and the hidden states from the encoder. This pre-training approach enables MASS to perform representation learning and language generation simultaneously. MASS has achieved significant improvements in several sequence-to-sequence tasks, such as neural machine translation and text summarization \cite{song2019mass}. 

Our augmentation method has a similar training objective as MASS, and includes a label-aware module to constrain the generation process. 

\section{Conditional Augmentation for ATE}
As mentioned before, we formulate the data augmentation of aspect term extraction (ATE) as a conditional generation task.~In this section, we first introduce the problem formulation, and then describe our augmentation method in detail.

\subsection{Problem Formulation}
Given a training set $\mathcal{D}$ of review texts, in which each sample includes a sequence of $n$ words $X=[x_1,x_2,...,x_n]$ and a label sequence $L=[l_1,l_2,...,l_{n}]$, where $l_i\in \{B,I,O\}$.~Here, $B$, $I$ and $O$ denote if a word is at the \emph{beginning}, \emph{inside} or \emph{outside} of an aspect term, respectively.~The objective of our augmentation task is to generate a new sentence consistent with $L$ and the aspect term.

\subsection{Our Approach}
The above augmentation is modeled as a fine-grained conditional language generation task implemented by a masked sequence-to-sequence generation model. As depicted in Figure \ref{framework}, the model adopts Transformer \cite{vaswani2017attention} as its basic architecture, and consists of a 6-layer encoder and a 6-layer decoder with 12 attention heads in each layer. The embedding size and hidden size are both 768, and the feed-forward filter size is 3072. The generation model is initialized with the pre-trained weights of MASS. To further incorporate the domain knowledge, we perform in-domain pre-training as in \cite{howard2018universal}.\footnote{The Amazon and Yelp review datasets are used as the Laptop and Restaurant domain corpora, respectively.}

\subsubsection{Training}\label{Training Stage}

The training process is illustrated in Algorithm \ref{algorithm1}. For each batch, we first sample a few examples from the training set with replacement (Line 4) according to a probability $p$ specified in Equation (\ref{sample_prob}). The chosen examples are then masked using the Fragment Masking Strategy function (Line 6) to generate training examples for our model. We elaborate on Algorithm \ref{algorithm1} in the following paragraphs.

\vspace{3mm}
\noindent\textbf{Fragment Masking Strategy}
\vspace{1mm}

\noindent The function $\emph{MaskFrag}$ (Line 6) is performed on the chosen examples to mask positions from $u$ to $v=u+r*\emph{length}(X)$, where $\emph{length}(X)$ is the length of sentence $X$. Each masking position is replaced by $\mathbb{[M]}$ only if its label is $O$. As a result, we obtain a partially masked sentence $\hat{X}$ and a fragment $Y=[y_1,y_2,...,y_{m}]=[x_u,x_{u+1},...,x_v]$, where $m=v-u+1$ is the length of the fragment.

\vspace{3mm}
\noindent\textbf{Sampling Strategy}
\vspace{1mm}

\noindent Line 5 of Algorithm \ref{algorithm1} shows that during the training process each sentence is masked every time it is sampled. Since long sentences have more different segments to mask than short ones, they should be sampled more frequently. We define the sampling probability $p_i$ of each example $i$ as follows:
    \begin{equation}
    \label{sample_prob}
    p_i\propto\left\{\begin{array}{rl}{\sqrt{d_{i}}\quad,} & {d_{i} > 5,} \\ 
    {0\quad,} & {otherwise}\end{array}\right.
    \end{equation}
where $d_i$ denotes the sequence length of example $i$.

\vspace{3mm}
\noindent\textbf{Training Objective}
\vspace{1mm}

\noindent The training objective (Line 9) takes the masked sentence $\hat{X}$ and label sequence $L$ as input, and reconstructs the masked fragment $Y$.~The inputs of the encoder are obtained by summing up the embeddings of a token $\hat{x}$, its aspect label ${l}$, and position ${q}$. The output is the hidden state $H=[h_1,h_2,...,h_n]$:
\begin{equation}\label{eq1}
H = {Enc}(\hat{X}, L),
\end{equation}
where $Enc$ represents the encoder, and $h_t \in \mathbb{R}^{s_{h}}$ denotes the hidden state of size $s_{h}$ for word $\hat{x_{t}}$.

\renewcommand{\algorithmicrequire}{ \textbf{Input:}}
    \begin{algorithm}[t] 
        \caption{ Training} \label{algorithm1}
        \begin{algorithmic}[1]
            \REQUIRE {\  \\ $\mathcal{D}$: training set;\\ $\theta$: model parameters;\\ $p$: sampling probability from $\mathcal{D}$;\\ $K$: total training iterations;\\ $B$: batch size; \\ $r$: proportion of sentence to mask; 0.5 by default;}
    
            \FOR {$i \leftarrow 1\ to\ K$}
             \STATE {$Batch^{(i)} \leftarrow \emptyset$};
             \FOR {$j \leftarrow 1\ to\ B$}
              \STATE {Randomly sample an example $(X, L )$ from $\mathcal{D}$ by probability $p$};
              \STATE {Randomly sample a start position $u$, where $u\in( 1, (1-r) * \emph{length}(X) )$ };
              \STATE { $\hat{X}, Y \leftarrow \emph{MaskFrag}(X, L, u, r)$};
              \STATE { $Batch^{(i)} \leftarrow Batch^{(i)} \cup \{ (\hat{X}, L, Y)\} $ };
             \ENDFOR
             \STATE {$ \theta \leftarrow \emph{Train}(Batch^{(i)}, \theta)$};
            \ENDFOR\\
            \STATE {$\textbf{Return}\ \theta$};
    
                \end{algorithmic}
    \end{algorithm}

Each self-attention head of the encoder learns a representation for the sentence based on tokens $\hat{X}$, label sequence $L$ and position $Q$. The objective of the decoder is to generate a sequence $Y$ based on $\hat{X}$ and $L$. In particular, it predicts next token $y_t$ based on context representations $H$, current aspect label $l_t$ and previous tokens $[y_1,...,y_{t-1}]$. 
\begin{equation}\label{eq2}
P(Y|X, L) = \prod_{t=1}^{m}P(y_{t}|y_{1:t-1}, l_t, H),
\end{equation}
where the conditional probability of token $y_t$ is defined by:
\begin{equation}\label{eq3}
P(y_{t}|y_{1:t-1}, l_t, H) = softmax(Ws_t+b).
\end{equation}
Here, $W\in \mathbb{R}^{|V| \times s_{h}}$, $|V|$ is the vocabulary size, and $s_t$ is the hidden state of the decoder at time step $t$, being calculated as:
\begin{equation}\label{eq4}
s_t = z_t + Emb_l(l_{t}),
\end{equation}
\vspace{-10pt}
\begin{equation}\label{eq5}
z_t = \emph{Dec}(x_{t-1}, l_{t-1}),
\end{equation}
where $Emb_l$ is the label embedding function and $\emph{Dec}$ is the decoder. 

In Equation (\ref{eq4}), each decoding step is conditioned on the context information and the whole label sequence, making the generation controllable.

The encoder and the decoder are jointly trained by maximizing the log-likelihood loss: 
\begin{equation}\label{eq6}
\mathcal{J}= \sum_{t=1}^{m}\log \left(P_{\theta}\left(y_{t}| y_{1: t-1}, l_t, H\right)\right),
\end{equation}
where $\theta$ includes the trainable parameters.

\subsubsection{Augmentation}
\label{Augmentation}
After training for a few epochs, our model is used to predict the words in a masked fragment. Specifically, given an example $(X,L)$ from the training set $\mathcal{D}$, we choose a start position $u$ and apply $\emph{{MaskFrag}}(X,L,u,r)$ to obtain $\hat{X}$. To avoid that same positions are chosen repeatedly, we manually choose the start position $u$ for the augmentation. At generation time, we use beam search with a size of 5 for the auto-regressive decoding. After the decoder produces all the tokens compatible with the original label sequence and aspect terms, we obtain a new example $(\tilde{X},L)$. Empirically, we find the model tends not to generate a same segment as the old one when the masked segment is longer than 4. The above process can be run multiple times with different start positions, and generates multiple new examples from a source example. In this approach, each source example is augmented in turn.

\section{Experiment}
In this section, we first introduce the experimental datasets and several popular ATE models. Then, we report the experimental results, which are obtained by averaging five runs with different initializations.

\subsection{Datasets}
Two widely-used datasets, the Laptop from SemEval 2014 Task 4 \cite{pontiki-etal-2014-semeval} and the Restaurants from SemEval 2016 Task 5 \cite{pontiki2016semeval}, are used for our evaluations. The statistics of the two datasets are shown in Table \ref{table1}, which tells clearly that there are only a limited number of samples in both datasets.

    \begin{table}[h]
    \centering
    \begin{tabular}{p{14.5mm}p{0.9cm}<{\centering}p{1.15cm}<{\centering}p{0.9cm}<{\centering}p{1.15cm}<{\centering}}
    \toprule
    \multirow{2}{*}{\textbf{Dataset}} & \multicolumn{2}{c}{\textbf{Training}} & \multicolumn{2}{c}{\textbf{Testing}} \\
    \cmidrule(lr){2-3} \cmidrule(lr){4-5}
     & \textbf{\#Sent} & \textbf{\#Aspect} & \textbf{\#Sent} & \textbf{\#Aspect}\\
    \midrule
    Laptop &3045&2358 &800&654\\
    Restaurant&2000&1743&676&622 \\
    \bottomrule
    \end{tabular}
    \caption{Statistics~of~our~datasets.~\#Sent and \#Aspect denote the count of sentence and aspect,~respectively.}\label{table1}
    \end{table}
    
    \begin{table*}[ht]
    \center
    \begin{tabular}{p{3.2cm}p{1.9cm}<{\centering}p{2.9cm}<{\centering}p{1.9cm}<{\centering}p{2.5cm}<{\centering}}
    \toprule
    \multirow{2}{*}{Model} & \multicolumn{2}{c}{Laptop} & \multicolumn{2}{c}{Restaurant} \\ 
    \cmidrule(lr){2-3}\cmidrule(lr){4-5}
     & source & augmentation & source & augmentation \\ 
     \midrule[1pt]
    BiLSTM-CRF & 73.42 & \textbf{74.28} & 69.16 & \textbf{71.44} \\
    Seq2Seq for ATE & 76.68 & \textbf{78.68} & 73.71 & \textbf{74.01} \\
    BERT-FTC & 79.39 & \textbf{81.14} & 74.75 & \textbf{75.89} \\
    DE-CNN & 81.08 & \textbf{81.58} & 74.52 & \textbf{75.19} \\ 
    BERT-PT & 84.59 & \textbf{85.33} & 77.49 & \textbf{80.29} \\ 
    \bottomrule
    \end{tabular}
    \caption{F1-score(\%) obtained on the tests for various models, where \emph{source} denotes the original datasets.}\label{table2}
    \end{table*}

\subsection{Experimental Setup}

\subsubsection{Dataset Augmentation}
\label{dataset}
For each of the two datasets, we hold out 150 examples from the original training set for validation. For each remaining training example, we generate four augmented sentences according to Section \ref{Augmentation} with the proportion $r$ set to 0.5. The four new sentences are allocated to four different sets. This leads to four generated datasets.

\subsubsection{ATE Models} 
\label{eval_models}
To examine our data augmentation method, we use the original training sets and the augmented training sets to train several ATE models.~The details of these models are as follows.

\textbf{BiLSTM-CRF} is a popular model for sequence labeling tasks. Its structure includes a BiLSTM followed by a CRF layer \cite{lafferty2001conditional}. The word embeddings for this model are initialized by GloVe-840B-300d \cite{pennington2014glove} and fixed during training. The hidden size is set to 300, and we use Adam \cite{Kingma2014adam} with a learning rate of 1e-4 and L2 weight decay of 1e-5 to optimize this model.

\textbf{Seq2Seq for ATE} \cite{ma2019exploring} is the first effort to apply a sequence-to-sequence model for aspect term extraction. It adopts GRU \cite{cho2014learning} for both the encoder and the decoder. The encoder takes a source sentence as input, and the decoder generates a label sequence as the result. This approach is also equipped with a gated unit network and a position-aware attention network.

\textbf{BERT for token classification} \cite{devlin2019bert} uses pre-trained BERT with a linear layer. We implement this model using open source\footnote{https://github.com/huggingface/transformers} and initialize its parameters with the pre-trained BERT-BASE-UNCASED model. We refer to this model as BERT-FTC in the following paragraphs.

\textbf{DE-CNN} \cite{xu2018double} uses two types of word embeddings: general-purpose and domain-specific embeddings.\footnote{https://howardhsu.github.io/}~While the former adopt GloVe-840B-300d, the latter are trained on a review corpus. They are concatenated and fed to a CNN model of 4 stacked convolutional layers.

\textbf{BERT-PT} \cite{xu-etal-2019-bert}\footnote{https://github.com/howardhsu/BERT-for-RRC-ABSA}  utilizes the weights of pre-trained BERT for initialization. To adapt to both domain knowledge and task-specific knowledge, it is then post-trained on a large-scale unsupervised domain dataset and a machine reading comprehension dataset \cite{rajpurkar-etal-2016-squad,rajpurkar-etal-2018-know}. So far, it is the state of the art for ATE.

The above models are all open-sourced and their default settings are employed in our experiments.

\subsection{Results and Analysis}
\subsubsection{Effect of Double Augmentation}
We combine the original training set with each of the four generated datasets (refer to \ref{dataset}) and obtain four augmented training sets, each doubling the original training set in size. For each model, we train it on the four augmented training sets, respectively, and take their average F1-scores on the test set. By comparing this score with the model trained on the original training set, we can examine if the augmented datasets improve the model.\footnote{The scores here may be slightly different from that in the original papers, but each pair of experiments in comparison were conducted strictly under the same setting.}

As shown in Table \ref{table2}, all the models are improved more or less based on the augmented datasets. Even for the sate-of-the-art DE-CNN and BERT-PT models, our augmentation also brings considerable improvements, which confirms that our augmentation approach can generate useful sentences for training a more powerful model for aspect term extraction.

\subsubsection{Effect of Multiple Augmentation}
\begin{figure}[!h]
    \centering
    \vspace{-0.1cm}
    \includegraphics[width=1\linewidth]{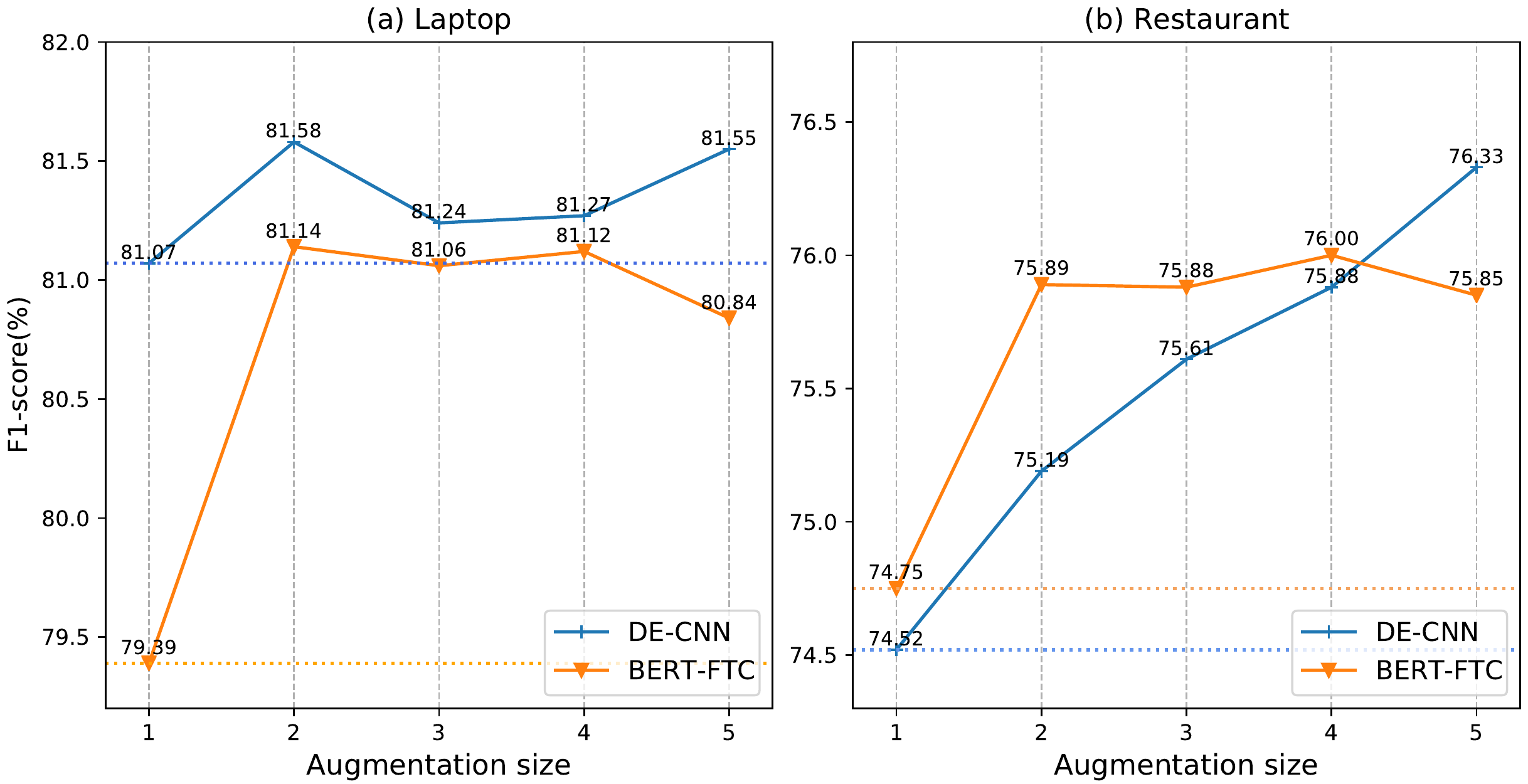}
    \caption{Performances of DE-CNN and BERT-FTC on different-sized augmentation datasets, where 1 means the original datasets without augmentation. All the results are based on the average scores of five runs.}
    \label{fig2}
    \end{figure}

The above results show the effect of double augmentation. In this subsection, we further combine any two of the four generated datasets with the original training set to form triple-augmented datasets, leading to six new datasets. In a similar approach, we can create quadruple-augmented and quintuple-augmented datasets. Then, we train the DE-CNN and BERT-FTC models on the new datasets and take the average F1-score for each model as before. The results are shown in Figure \ref{fig2}.

We can observe from the figure that both models are generally improved as the size of augmentation increases on the Restaurant dataset. There is even a 1.8 boost for DE-CNN. On the Laptop dataset, however, the highest scores are seen at double-augmentation for both models. One of the reasons could be the relatively large volume of the original dataset. Another possible reason is that the aspect terms in this dataset are often regular nouns such as \textit{screen} and \textit{keyboard}, which can be successfully extracted just based on their own meanings. Differently, aspect terms in the Restaurant dataset are more arbitrary and diverse such as \textit{Cafe Spice} and \textit{Red Eye Grill}, the names of dish or restaurant. This requires a model to pay more attention to the contexts while determining whether the candidate is an aspect terms. As our augmentation approach can generate different contexts for an aspect term, it works better on the Restaurant dataset. 

\section{Discussion}
In this section, we present more qualitative analysis and discussions about our augmentation approach.

\subsection{Does Larger Masked Length Help?}

In the augmentation stage, the masked proportion $r$ is a hyperparameter and set to the half of the length of a sentence in the above experiments. In this subsection, we explore its influence by changing it from 30\% to 70\% of sentence length stepped by 10\%. We use DE-CNN model for this evaluation on the double-augmented datasets.

\begin{figure}[h]
    \centering
    \includegraphics[width=1\linewidth]{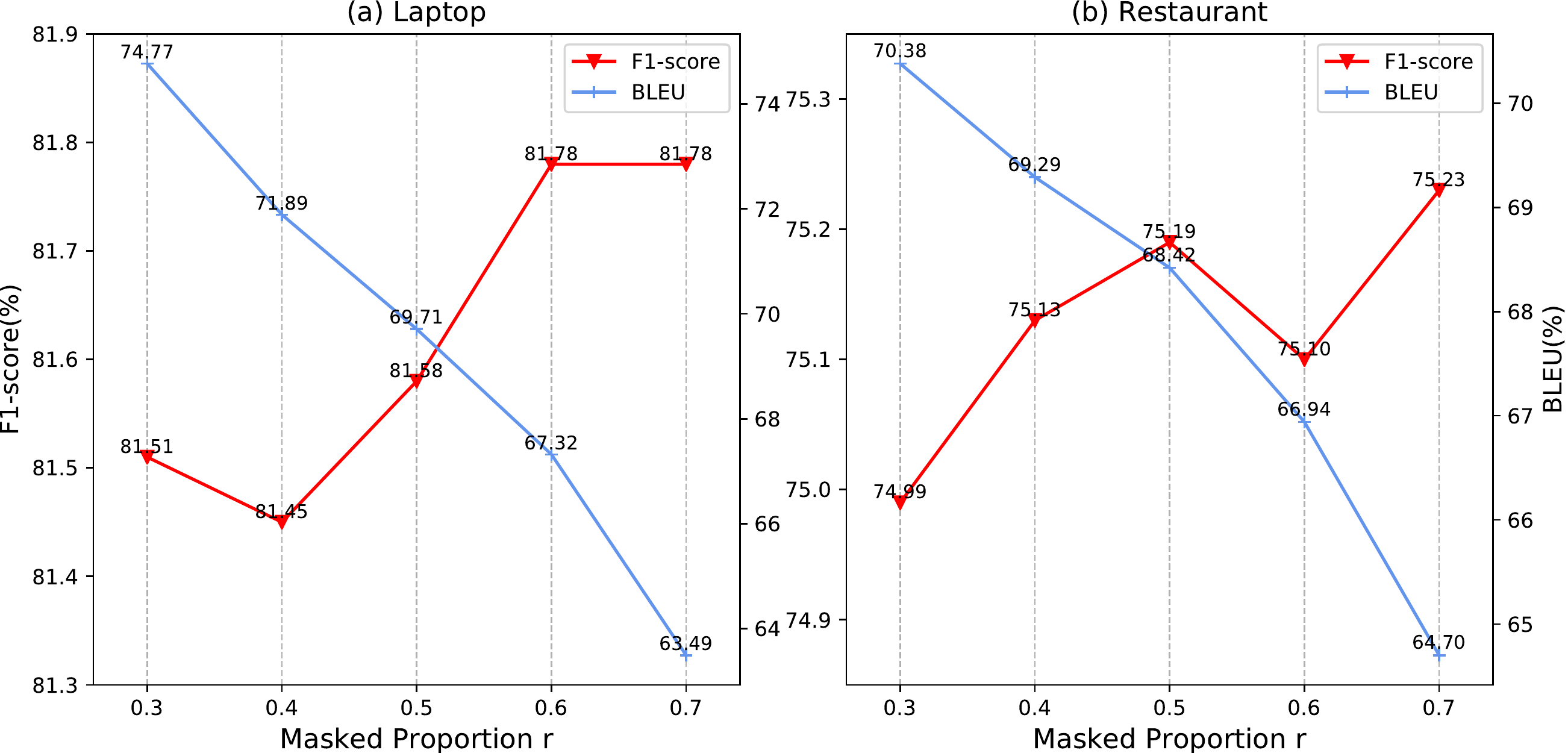}
    \caption{Performance of DE-CNN with different masked proportion $r$ for augmentation. BLEU is used to measure the correspondence between a new sentence after augmentation and the original sentence.}
    \label{fig3}
    \end{figure}

As shown in Figure \ref{fig3}, the overall trend for F1-scores is moving up as $r$ increases. The reason is that sentences with short masked fragments are likely to be restored to their original forms by our generation model. As the proportion $r$ increases, the content of a sentence has increasingly more chances to be changed significantly, resulting in diversified new sentences. This can be confirmed by the declining BLEU scores in Figure \ref{fig3}.

\begin{table*}[!t]
    \center
            \resizebox{2\columnwidth}{!}{%
    \begin{tabular}{p{2.6cm}p{1.6cm}<{\centering}p{1.6cm}<{\centering}p{1.6cm}<{\centering}p{1.6cm}<{\centering}p{1.6cm}<{\centering}p{1.6cm}<{\centering}}
    \toprule
    \multirow{2}{*}{\textbf{Dataset}} & \multicolumn{3}{c}{\textbf{Laptop}} & \multicolumn{3}{c}{\textbf{Restaurant}} \\ 
   \cmidrule(lr){2-4} \cmidrule(lr){5-7} 
     & \textbf{Precision} & \textbf{Recall} & \multicolumn{1}{c}{\textbf{F1}} & \textbf{Precision} & \textbf{Recall} & \multicolumn{1}{c}{\textbf{F1}} \\ 
     \midrule
    Source & 81.24 & 80.91 & 81.08 & 70.62 & 78.88 & 74.52 \\ 
    Ours w/o LEM & 80.75 & 79.66 & 80.20 & 70.63 & 78.23 & 74.24 \\ 
    \rowcolor{gray!20} Ours & \textbf{81.88} & \textbf{81.29} & \textbf{81.58} & \textbf{70.86} & \textbf{80.08} & \textbf{75.19} \\ 
    \bottomrule
    \end{tabular}
    }
    \caption{Results of ablation study on whether label embeddings are used, where \emph{Source} denotes the original dataset, and \emph{Ours w/o LEM} denotes our augmentation model without label embeddings.}
    \label{table3}
    \end{table*}

\subsection{Does Label Sequence Matter?}

Our augmentation model introduces label embeddings into Transformer to force the new sentences to be task-competent.~We conduct an ablation study to verify the effectiveness by removing these embeddings during augmentation.~The DE-CNN model is used again for this study.

As shown in Table \ref{table3}, the removal of label embeddings causes considerable performance drops, and the results are even worse than that on the original dataset. This is probably due to the poor \emph{Recall} performance that can be explained as follows. When label sequence information is not present, the augmentation is prone to produce decayed examples in which some new aspect terms are generated in the positions of label $O$, or verse vice. The model trained with such decayed examples is misled not to extract these aspect terms in the test stage. As a result, the model makes many false-negative errors, leading to poor \emph{Recall} scores. This indicates that label embeddings are helpful for generating qualified sentences for aspect term extraction.

 \begin{table*}[!t]
    \center
        \resizebox{2.05\columnwidth}{!}{%
    \begin{tabular}{p{2.5cm}p{3.3cm}<{\centering}p{1.2cm}<{\centering}p{1cm}<{\centering}p{3.3cm}<{\centering}p{1.2cm}<{\centering}p{1.1cm}<{\centering}}
    \toprule
    \multirow{2}{*}{\textbf{Dataset}} & \multicolumn{3}{c}{\textbf{Laptop}} & \multicolumn{3}{c}{\textbf{Restaurant}} \\ 
          \cmidrule(lr){2-4} \cmidrule(lr){5-7} 
& \textbf{Precision/Recall/F1} & \textbf{BLEU} & \textbf{Fluency} & \textbf{Precision/Recall/F1} & \textbf{BLEU} & \textbf{Fluency} \\ 
    \midrule
    {Source} & 77.36/81.53/79.39 & 100.00 & 203 & 74.86/74.64/74.75 & 100.00 & 158 \\ 
    {DATA\_synonym} & 74.91/79.06/76.93 & 76.68 & 558 & 71.23/76.43/73.74 & 78.22 & 438 \\
    {DATA\_BERT} & 80.06/78.60/79.32 & 69.93 & 461 & 72.34/76.37/74.30 & 68.57 & 435 \\ 
    {DATA\_GPT-2} & 71.97/61.10/66.09 & \textbf{54.86} & 328 & 70.53/57.42/63.30 & \textbf{58.60} & 314 \\ 
   \rowcolor{gray!20} {DATA\_Ours} & 80.27/80.98/\textbf{80.61} & 69.70 & \textbf{242} & 73.95/76.95/\textbf{75.41} & 68.42 & \textbf{257} \\ 
    \bottomrule
    \end{tabular}
    }
    \caption{Results on datasets generated by different augmentation approaches. \emph{Source} denotes the original datasets; \emph{DATA\_synonym} denotes the datasets obtained by randomly replacing tokens with their synonyms. \emph{DATA\_BERT}, \emph{DATA\_GPT-2} and \emph{DATA\_Ours} denote the datasets generated by BERT, GPT-2 and our augmentation approach, respectively.~BERT-FTC is used as the implementation model, and lower \emph{Fluency} scores mean more fluent sentences.}
    \label{table4}
    \end{table*}

\begin{table*}[!t]
    \center
    \resizebox{2.05\columnwidth}{!}{%
    \begin{tabular}{lp{14.5cm}}
    \toprule
    \textbf{Source:} & Also, \textcolor{blue}{the} \textbf{space bar} \textcolor{blue}{makes a noisy click} every time you use it.\\
     \textbf{Augmented:} & Also, \textcolor{purple}{the} \textbf{space bar} \textcolor{purple}{will get stuck there} every time you use it.\\
        \midrule
            \textbf{Source:} & The \textbf{hinge design} \textcolor{blue}{forced you to place various connections all} around the computer, left right ...\\
     \textbf{Augmented:} & The \textbf{hinge design }\textcolor{purple}{also allows you to adjust the angle} around the computer , left right ...\\
        \midrule
            \textbf{Source:} & Their \textbf{pad penang} \textcolor{blue}{is delicious and} everything else is fantastic.\\
     \textbf{Augmented:} & Their \textbf{pad penang} \textcolor{purple}{is mediocre but} everything else is fantastic.\\
        \midrule
            \textbf{Source:} & \textcolor{blue}{I am learning the finger options for the} \textbf{mousepad} that allow for quicker browsing of web pages.\\
     \textbf{Augmented:} &  \textcolor{purple}{I also enjoy the fact that it has a} \textbf{mousepad} that allow for quicker browsing of web pages.\\
        \midrule
            \textbf{Source:} & \textcolor{blue}{I charge it at night and skip taking the} \textbf{cord} with me because of the good \textbf{battery life}. \\
     \textbf{Augmented:} & \textcolor{purple}{I don't have to carry the} \textbf{cord} with me because of the good \textbf{battery life}. \\
    \bottomrule
    \end{tabular}
    }
    \caption{Examples generated by our augmentation approach. Texts in bold, blue and purple represent aspect terms, masked fragments and generated fragments, respectively. }
    \label{case}
    \end{table*}

\subsection{Why Sequence-to-Sequence Generation?}
As mentioned before, we formulate the data augmentation for aspect term extraction as a conditional generation problem that is solved by masked sequence-to-sequence learning. One may argue that other pre-trained language models like BERT and GPT-2 are also competent for this task as in \cite{DBLP:conf/iccS/WuLZHH19, sudhakar-etal-2019-transforming, keskar2019ctrl}. Here we compare them and demonstrate the superiority of our approach in this task.
    
Following some previous work \cite{DBLP:conf/iccS/WuLZHH19,sudhakar-etal-2019-transforming,keskar2019ctrl}, we modify the settings of BERT and GPT-2 to make them fit this task. Readers are recommended to refer to Appendix for more details. Moreover, a widely-used replacement-based method is implemented for comparison, in which half of the tokens are randomly replaced by their synonyms from WordNet \cite{miller1995wordnet}. We use \emph{fluency}\footnote{Fluency is measured by the \emph{perplexity} of sentence, and is calculated by OpenAI GPT. In this metric, sentences with lower perplexity scores are more fluent. Note that the \emph{GPT} here is different from GPT-2 that we use to generate text data.} and BLEU\footnote{The original sentences are taken as reference.} to evaluate the generated sentences. Note that these datasets do not contain the original training examples because we want to focus more on the generated ones. We employ BERT-FTC as the implementation model and train it on these datasets. The results on the test sets are presented in Table \ref{table4}.

From the table, we note that the F1 scores of GPT-2 are the worst because of its low recall scores. This conforms with the architecture and the language modeling objective of GPT-2, which does not have an encoder to encode the label information. In this case, the decoding step is uncontrollable and cannot generate a sentence fitting the label sequence. In contrast, our framework contains an encoder to encode a sentence and the label sequence simultaneously, and a decoder to generate sentences conditional on the encoder output. That is, our decoder takes advantage of both context information and aspect label information, making the augmentation conditional and controllable.
 
BERT performs the worst in this task in fluency. This can be attributed to its independence assumption in the process of generation, which means that all masked tokens are independently reconstructed, likely leading to in-coherent word sequences. In contrast, our approach generates the sequence in an auto-regressive way, with each decoding step based on the result of its previous step, ensuring fluent new sentences.

The replacement-based method does not take into account the sentence context and leads to poor fluency scores. Also, there are limited words to choose for synonyms in such lexical databases as WordNet. Thus, such replacement-based methods can only produce sentences of limited diversity, which is confirmed by the BLEU scores.  

To sum up, our data augmentation model benefits considerably from its encoder-decoder architecture and the masked sequence-to-sequence generation mechanism, which is controllable to ensure qualified data augmentation for aspect term extraction. The results show that this sequence-to-sequence generation framework is non-replaceable by other language models such as BERT and GPT-2.

\vspace{-0.051cm}
\subsection{Case Study}
We finally present several augmented examples in Table \ref{case} to illustrate the effect of our augmentation method more intuitively. We observe that the contents of the masked fragments can be dramatically changed from their original forms after augmentation. In some cases, the sentiment polarities are even reversed. Nevertheless, the new contexts are still appropriate for the aspect terms, making them qualified and also diversified new training examples for aspect term extraction.

\section{Conclusion}
In this paper, we have presented a conditional data augmentation approach for aspect term extraction. We formulated it as a conditional generation problem and proposed a masked sequence-to-sequence generation model to implement it. Unlike existing augmentation approaches, ours is controllable to generate qualified sentences,~and allows more diversified new sentences. Experimental results on two review datasets confirm its effectiveness in this conditional augmentation scenario. We also conducted qualitative studies to analyze how this augmentation approach works, and tested other language models to explain why our masked sequence-to-sequence generation framework is favored. Moreover, the proposed augmentation method tends not to be unique to the current task and could be applied to other low-resource sequence labeling tasks such as chunking and named entity recognition.

\section*{Acknowledgments}
The work was partially supported by the Fundamental Research Funds for the Central Universities (No.19lgpy220) and the Program for Guangdong Introducing Innovative and Entrepreneurial Teams (No.2017ZT07X355). 

\bibliography{anthology,acl2020}
\bibliographystyle{acl_natbib}

\newpage
\appendix

\section{Appendices}
\label{sec:appendix}

We modify the settings of BERT and GPT-2 to make them fit our augmentation task as follows.
\\ \hspace*{\fill} \\
\textbf{BERT} We follow \citet{DBLP:conf/iccS/WuLZHH19} who applied the masked language model task to augment their training data for text classification. The segment embeddings in the input layer are replaced by label embeddings. 
The objective is to predict the masked token $t_{i}$ based on the conditional probability distribution $p\left(t_i | L, \boldsymbol{S} \backslash\left\{t_{i}\right\}\right)$, where $\boldsymbol{S}$ and $L$ denote the sentence and its label sequence, respectively, and $\boldsymbol{S} \backslash\left\{t_{i}\right\}$ means the context of $t_i$. To predict $t_i$, both the context $\boldsymbol{S} \backslash\left\{t_{i}\right\}$ and the label sequence $L$ are considered. The word with the highest probability among all vocabulary words is chosen.

In our experiments, 50\% of the tokens in a sentence are masked individually.~These tokens are then reconstructed in the masked language modeling task. Finally, the predicted tokens and the unmasked tokens constitute an augmented sentence.

\vspace{0.5cm}
\textbf{GPT-2} We refer to the work of \citet{sudhakar-etal-2019-transforming} and \citet{keskar2019ctrl}, which aim at style transfer and controllable generation, respectively. In the training stage, the format of input is: 
$\mathbb{[ASP]}$ $\textit{aspect}_1$ $\mathbb{[ASP]}$ $\textit{aspect}_2$ $\mathbb{[BOS]}$ $\textit{sentence}$ $\mathbb{[EOS]}$,
where $\mathbb{[ASP]}$ is a special symbol followed by an aspect term, and $\mathbb{[BOS]}$ is appended before the source sentence. For example, for a source sentence ``\textit{I was disappointed with this restaurant}", where ``\textit{restaurant}" is the aspect term, the input takes the following format: $\mathbb{[ASP]}$ $\textit{restaurant}$ $\mathbb{[BOS]}$ \textit{I was disappointed with this restaurant.} $\mathbb{[EOS]}$. Let $\boldsymbol{S}$ denote the source sentence. The auto-regressive model learns to reconstruct $\boldsymbol{S}$ given $C_{X}$, which is the fragment of the input from the first token $\mathbb{[ASP]}$ to $\mathbb{[BOS]}$. The objective is to maximize:
\begin{equation}
    L(\theta)= \log p\left(\boldsymbol{S}| C_{X}; \theta\right)
\end{equation}
where $\theta$ denotes the parameters of GPT-2.

In the augmentation stage, this model takes the fragment $C_{X}$ and half of the source sentence as input, and tries to reconstruct the other half. An augmented sentence is formed by joining the unmasked half of the sentence and the reconstructed half. The beam size is set to 5 in our experiments.

\end{document}